\pdfoutput=1
\documentclass[11pt]{article}

\usepackage{ACL2023}

\usepackage{times}
\usepackage{latexsym}

\usepackage[T1]{fontenc}
\usepackage[utf8]{inputenc}

\usepackage{microtype}

\usepackage{inconsolata}

\usepackage{graphicx}
\usepackage[capitalise]{cleveref}
\usepackage{subcaption}
\usepackage{multirow}
\usepackage[colorinlistoftodos]{todonotes}
\usepackage{booktabs}
\usepackage{xurl}
\usepackage{marvosym}

\title{CoCoHD: Congress Committee Hearing Dataset}

\author{\hypersetup{linkcolor=black} Arnav Hiray$^{\spadesuit}$, Yunsong Liu$^{\spadesuit}$, Mingxiao Song$^{\spadesuit}$, Agam Shah\textsuperscript{\Letter}, Sudheer Chava\\
        Georgia Institute of Technology 
}

\begin{document}

\maketitle
%\\ \quad $\spadesuit$ Equal 
\def\thefootnote{\Letter}\footnotetext{Corresponding Author: \href{mailto:ashah482@gatech.edu}{ashah482@gatech.edu}}\def\thefootnote{\arabic{footnote}}
\def\thefootnote{$\spadesuit$}\footnotetext{Equal Contribution}\def\thefootnote{\arabic{footnote}}
%\footnote{\\ $\spadesuit$ Equal Contribution}

\begin{abstract}
  U.S. congressional hearings significantly influence the national economy and social fabric, impacting individual lives. Despite their importance, there is a lack of comprehensive datasets for analyzing these discourses. To address this, we propose the \textbf{Co}ngress \textbf{Co}mmittee \textbf{H}earing \textbf{D}ataset (CoCoHD), covering hearings from 1997 to 2024 across 86 committees, with 32,697 records. This dataset enables researchers to study policy language on critical issues like healthcare, LGBTQ+ rights, and climate justice. We demonstrate its potential with a case study on 1,000 energy-related sentences, analyzing the Energy and Commerce Committee's stance on fossil fuel consumption. By fine-tuning pre-trained language models, we create energy-relevant measures for each hearing. Our market analysis shows that natural language analysis using CoCoHD can predict and highlight trends in the energy sector.\footnote{Our dataset and code is available at \url{https://github.com/gtfintechlab/CoCoHD}.}
\end{abstract}

\begin{figure}[!ht]
  \centering
  \includegraphics[width=0.47\textwidth]{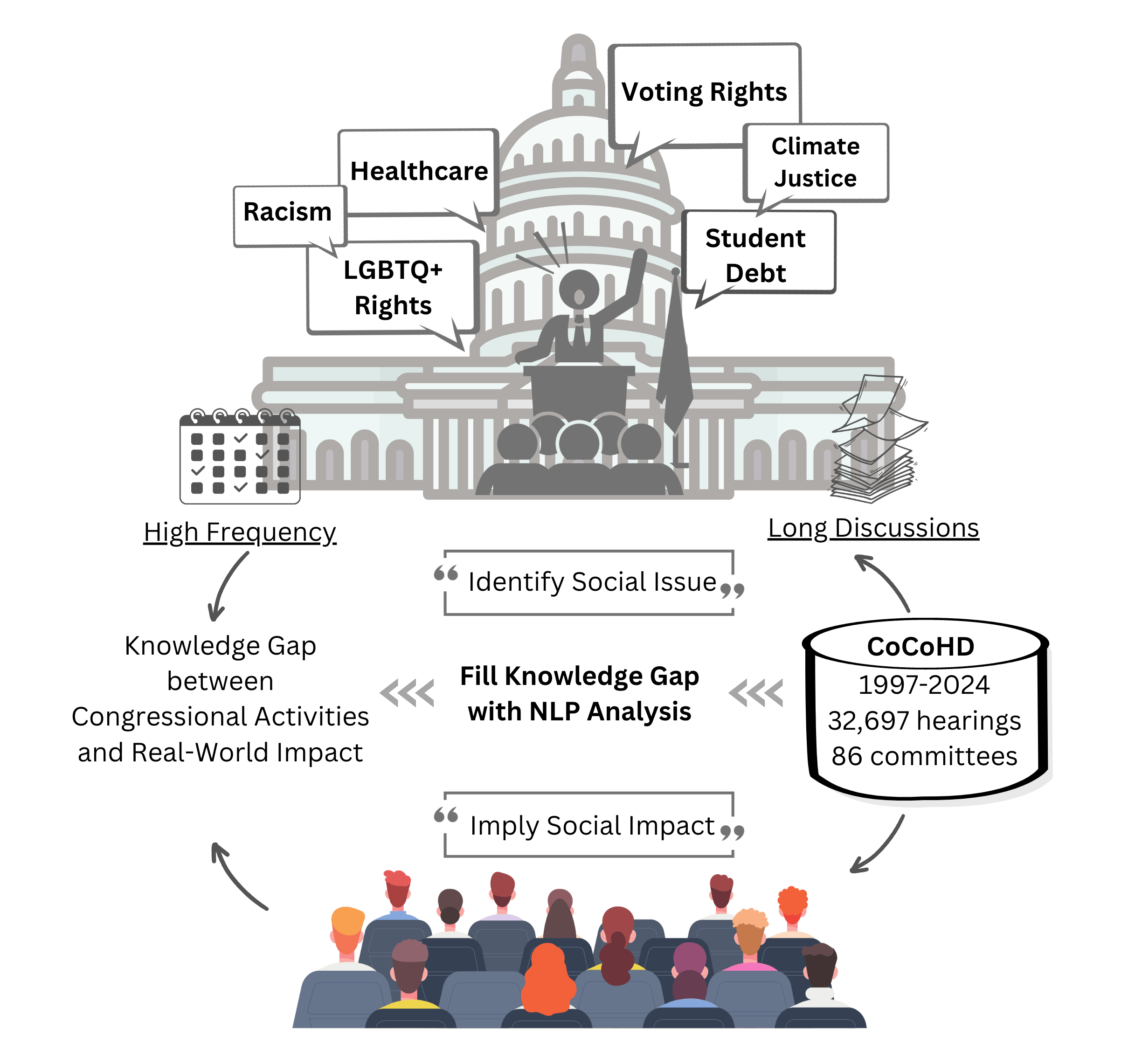}
  \caption{Visual illustration of the social impact of the CoCoHD dataset.}
  \label{fig:illustration}
\end{figure}

\section{Introduction}

The United States Congress, with its profound influence on the daily lives of millions of Americans and global resonance, consists of two chambers: the House of Representatives and the Senate. The House of Representatives has 435 members, with the number from each state determined by population, and members serve two-year terms. The House is responsible for initiating revenue bills and has the power to impeach federal officials. The Senate, comprising 100 members with two from each state regardless of population, serves six-year terms and provides a more deliberative body. The Senate has the authority to confirm presidential appointments, ratify treaties, and conduct impeachment trials. Both chambers must pass a bill in identical form for it to become law, ensuring thorough scrutiny and representation of both local and statewide interests in the legislative process. 

In the mid-20th century, congressional hearings became crucial tools for both political parties to advance their agendas. Congressional hearings are formal sessions where legislators gather evidence, question witnesses, and review government actions to inform policy decisions and ensure accountability. The advent of television and the Internet further transformed these events into political theater, engaging wider audiences and fostering debates on significant issues important to the country.\footnote{\url{https://www.nationalaffairs.com/publications/detail/reclaiming-the-congressional-hearing}}

We see this evident in the discussions on climate change. In 2023, the 118th United States Congress held 156 hearings on climate, environment, and energy topics.\footnote{\url{https://www.eesi.org/articles/view/on-the-hill-in-2023-a-breakdown-of-climate-energy-and-environmental-congressional-hearings}} One of the most debated subjects was the Inflation Reduction Act (IRA), designed to make renewable energy and clean technologies more affordable. The IRA has the potential to create over 9 million jobs in the next decade and has catalyzed over \$49 billion in investment in clean energy technologies since August 2022.\footnote{\url{https://www.bluegreenalliance.org/site/9-million-good-jobs-from-climate-action-the-inflation-reduction-act/}} These facts underscore Congress's extensive influence on economic and social dynamics and the evident impact of legislation discussed in congressional hearings.

The increased accessibility of congressional discussions has transformed the tone and format of hearings. Politicians now aim to persuade not only those present but also remote audiences, enhancing public engagement in the legislative process.\footnote{\url{https://www.washingtonpost.com/news/in-theory/wp/2016/09/30/why-congressional-hearings-still-matter/}} As socioeconomic challenges grow more complex, hearings have become larger, longer, and more intricate. This complexity makes it difficult to fully track their development, leading to a significant knowledge gap between congressional activities and public understanding. Bridging this gap requires resources and methods for large-scale analysis of the trends and issues discussed in these hearings. A comprehensive dataset and data-driven analyses will enable researchers, policymakers, and the public to systematically study congressional hearings, enhancing transparency, fostering informed discourse, and facilitating evidence-based decision-making on key legislative matters.
This paper aims to promote research and analysis in natural language processing (NLP) applications for congressional hearings by introducing CoCoHD (Congress Committee Hearing Dataset), the first and largest open-source dataset of its kind. CoCoHD comprises 32,697 curated hearing transcripts and metadata from 1997 to 2024, covering 86 congressional committees.

As a demonstration, we present a novel task for quantifying fossil fuel-related sentiment in congressional hearings, focusing on the Energy and Commerce Committee. We manually annotated sentences as relevant or irrelevant to energy production and further categorized relevant sentences as supportive, oppositional, or neutral toward fossil fuel production. By fine-tuning pre-trained language models with these annotations, we generalized this labeling process to all hearings from the committee, creating an inclination measure that quantifies each hearing's stance on fossil fuels versus clean energy. 

Our subsequent market analysis showed that this measure could explain and predict trends in the energy sector, highlighting CoCoHD's potential to bridge the gap between congressional hearings and their real-world impact. Beyond the energy sector, CoCoHD empowers researchers to explore language complexities in policy discussions on a wide range of critical issues, including immigration, climate change, racial justice, and LGBTQ rights.

\section{CoCoHD Dataset}

The Congress Committee Hearing Dataset contains details for 32,697 U.S. Congress hearings held by the United States Congress between January 1997 and January 2024, along with transcripts for 32,435 of those hearings. Accompanied with each hearing is metadata pertaining to each hearing. In this section, we explain and describe the dataset's structure and our data collection process.

\subsection{Current Limitations in Publicly Available Congressional Hearing Data}
GovInfo \cite{govinfo} is a service of the United States Government Publishing Office (GPO), which provides free public access to official publications from all three branches of the Federal Government \footnote{\url{https://www.govinfo.gov/}}. In particular, information and transcripts of Congressional Hearings can also be found. While GovInfo has made considerable effort to ensure the public availability of information discussed in Congressional Hearings, the currently available transcripts and associated  ``Content Details'' (hereby referred to as Metadata) contain inherent limitations that hinder the widespread usage and accessibility of these transcripts. 

\paragraph{Inconsistent Committee Naming and and Missing Subcommittee Information}

One of the notable limitations of the current raw congressional hearing transcripts is the inconsistent naming of committees. For example, typos are common, and some committees are inconsistently referred to, such as "Committee on Oversight and Government Reform" and "Committee on Government Reform and Oversight," posing a difficulty in accurately tracking and analyzing grouped data based on the transcripts and associated metadata. In addition to this, the existing metadata available via GovInfo does not contain details regarding the subcommittees involved, despite saying so.

\paragraph{Unclear Subdivisions of Sections} The raw text of hearing transcripts contains various, non-standardized patterns and formatting, indicating subdivisions within the transcript. Phrases in full capitalization signal titles or provide information about committees, subcommittees, speakers, and witnesses. However, the lack of standardization leads to inconsistencies, such as missing dividers, inconsistent indentations and line breaks, and erratic capitalization. Text that should be capitalized often isn’t, and phrases can be split across multiple lines with varying spacing. These formatting issues complicate the use of regular expressions to reliably split transcripts by paragraphs, sentences, sections, and speakers.

\begin{table}[ht]
\centering
\footnotesize
\begin{tabular}{@{}ll@{}}
\toprule
\multicolumn{1}{c}{\textbf{GovInfo ID}} & Unique ID of the hearing. \\
\multicolumn{1}{c}{\textbf{Title}} & Hearing title. \\
\multicolumn{1}{c}{\textbf{Held Date}} & Hearing date. \\
\multicolumn{1}{c}{\textbf{Congress}} & Congress iteration. \\
\multicolumn{1}{c}{\textbf{Congress Chamber}} & House, Senate, or Joint. \\
\multicolumn{1}{c}{\textbf{Committee}} & Hearing committee. \\
\multicolumn{1}{c}{\textbf{Mapped Committee}} & Standardized committee name. \\
\multicolumn{1}{c}{\textbf{Members}} & Members listed. \\
\multicolumn{1}{c}{\textbf{Witnesses}} & Listed witnesses. \\
\multicolumn{1}{c}{\textbf{Serial Numbers}} & Sequence \# within committee. \\
\multicolumn{1}{c}{\textbf{Bill Numbers}} & Referenced bill numbers. \\
\multicolumn{1}{c}{\textbf{Hearing Content}} & Table of contents. \\
\multicolumn{1}{c}{\textbf{Prepared Statements}} & Associated prepared statements. \\
\bottomrule
\end{tabular}
\caption{Brief description of the metadata contents for each hearing in the dataset.}
\label{tab:metadata_table}
\end{table}

\subsection{Dataset Construction}

\begin{figure}[ht]
  \centering
  \includegraphics[scale = .5]{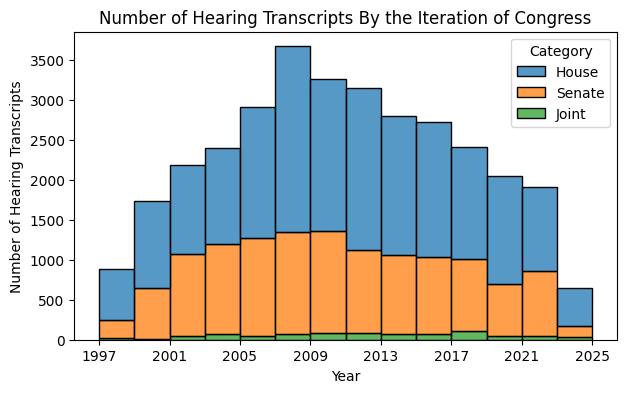}
  \caption{Hearing transcript counts by Congress years.}
  \label{fig:transcripts_by_year}
\end{figure}

Transcripts in CoCoHD all follow the same structure. The majority of a transcript consists of speech by committee members and witnesses. Additionally, each transcript provides a title page and a table of contents that lists speakers from the committee and witnesses with their backgrounds. A total of 32,435 transcripts are collected. On average, each transcript has 33,274 words. These congressional hearings are split into hearings led by the House of Representatives, the Senate, and Joint Hearings as shown in Figure~\ref{fig:transcripts_by_year}. CoCoHD also provides a metadata file for 32,697 hearings. 

\paragraph{Congressional Hearing Transcripts} While congressional committees are not required to publish transcripts of hearings, the transcripts of most hearings are now distributed in text format on GovInfo, a website maintained by the U.S. Government Publishing Office. It is important to note that it may take months to even years for transcripts to be publicly available on GovInfo. To transform fragmented hearing data from the website into an NLP dataset, we scraped the hearing details and transcripts, designed an easy-to-use dataset structure, filtered out erroneous transcripts, and categorized hearings by committee.

We crawled hearing details and transcripts in a two-stage process due to the lack of an API. In the first stage, a list of hearings, with links to each hearing's transcript and details page, was obtained. Subsequently, transcript files were downloaded, and hearing details were scraped and stored as one list in JSON format.

\subsection{Metadata}
Metadata is essential for organizing and understanding complex datasets. In the context of congressional hearings, metadata provides crucial details such as participant names, dates, and topics discussed, enhancing data accessibility and analysis. This detailed context allows researchers to efficiently locate relevant information, ensuring accurate and reproducible findings in legislative research. We believe it is crucial to enhance the metadata provided as this is uniquely key in order to better identify and understand topics of interest. Table \ref{tab:metadata_table} provides a brief description of sections we retrieved from GovInfo. In addition to this information, we provided new information suited to better understanding congressional hearings.  

\paragraph{Standardizing Committees and Identifying Subcommittees} To address issues surrounding the inconsistent labeling of committees in the content detail section on GovInfo, we identified each variation present in the metadata and created an appropriate dictionary matching them to the correct committee names. This dictionary is available as a JSON file in our dataset. We also update the metadata accordingly. Furthermore, we also identify the subcommittees involved in 17,507 hearings. We explain the method of identification in Appendix~\ref{sec:identify_subcommittees_ap}.

\paragraph{Identifying Content of Hearing}

Many pre-planned congressional hearings contain a subsection detailing the contents of the hearing. This may provide information on the planned events such as speakers and submissions of prepared statements. The Contents section also provides information on witnesses. We have extracted the contents of hearings when available as a complement to information regarding witnesses and members present that is already provided by GovInfo. We detail the method of identification in Appendix~\ref{sec:hearing_contents_ap}. 

\paragraph{Identifying Prepared Statements}
While information surrounding members and other details regarding content are available, there currently is not a method for uniquely identifying Prepared Speeches or Prepared Letters to be presented during Congressional Hearings. As such, we have uniquely identified the speakers/authors of prepared statements (including opening statements) in the metadata. We detail the method of identification in Appendix~\ref{sec:prepared_statements_ap}.

\paragraph{Tracking Speaker Participation and Frequency}
In addition to witnesses speaking at congressional hearings, often members of Congress discuss topics and question witnesses. To highlight this, we identify members present at the congressional hearing who engage in dialogue. We retrieve their names and the number of times they speak. We detail the method of identification in  Appendix~\ref{sec:active_speakers_ap}.

\begin{figure*}[ht]
  \centering
  \includegraphics[width=\textwidth]{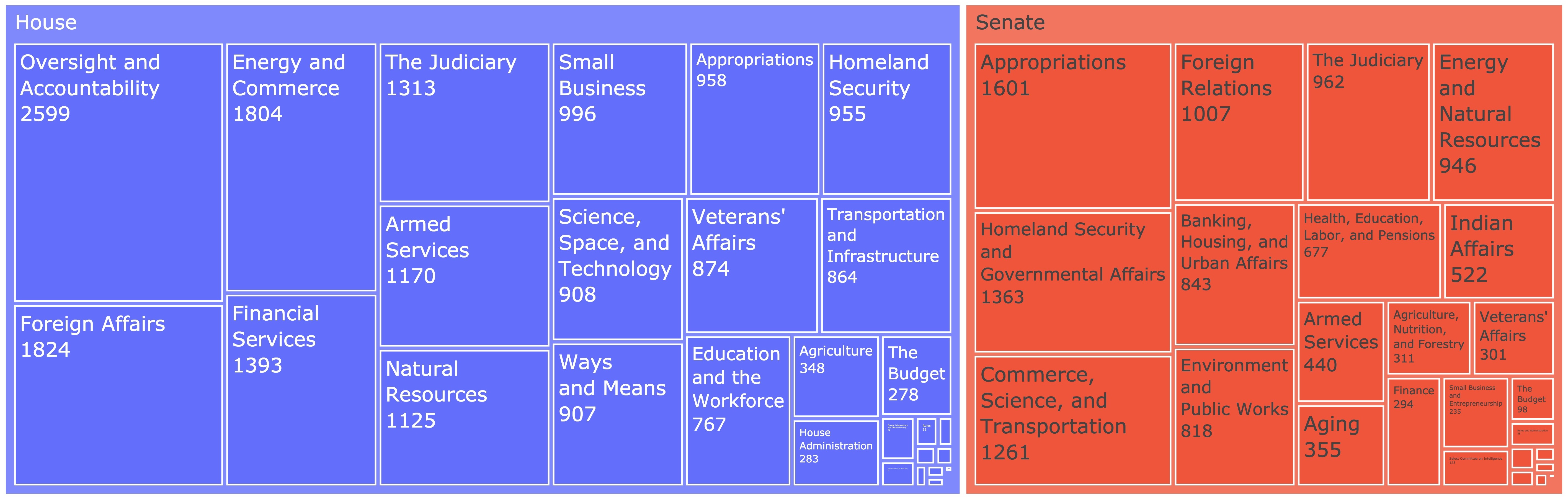}
  \caption{Distribution of hearings across committees and chambers, excluding joint hearings}
  \label{fig:tree_map}
\end{figure*}

\paragraph{Statistics}
sOn average, a hearing transcript in the CoCoHD dataset contains 1,810 sentences and 38,724 words. The number of words per sentence is approximately 21.38. During 1997-2023, house committees published transcripts for 19,606 hearings, and Senate committees published 12,299 transcripts. Figure \ref{fig:tree_map} demonstrates that the three most active committees are Oversight and Accountability, Foreign Affairs, and Energy and Commerce. In comparison, committees on Appropriations and Homeland Security and Governmental Affairs are the most active Senate committees. 

\paragraph{Open/Source Public Release }
To the best of our knowledge, our dataset is the first collection of Congressional Hearings pre-processed and does not require scraping from Govinfo. Furthermore, publicly releasing the code for scraping hearings and transcript-level filtering will enable researchers to quickly gather transcripts in the future. This will enable accessibility for a better understanding of the language used by politicians and their impact.

There are many types of tokenization that can be done before utilizing a textual dataset, such as word tokenization, subword tokenization, character tokenization, sentence tokenization, n-gram tokenization, Byte-Pair Encoding (BPE), WordPiece tokenization, whitespace tokenization, punctuation-based tokenization, and regex tokenization. As different tasks may prefer or require different tokenization methods, we have intentionally decided not to tokenize the dataset but to leave it in the raw text. This can allow for diverse usage and analysis.

\section{Applications}
Our congressional hearing dataset is enriched with a diverse set of metadata, providing researchers the flexibility to easily combine multiple sources and tailor their analyses to specific research goals. In the following application section, we introduce three levels of analysis that can be conducted with our dataset. By cross-referencing existing literature, we demonstrate that our dataset is capable of satisfying various research needs. Finally, we present a working example case study, illustrating how the dataset can be utilized to quantify Congress' inclination on energy policies.

\subsection{Perspectives}
The CoCoHD dataset can be further processed and analyzed with NLP techniques from three perspectives: inter-hearing, intra-hearing, and participant. For each perspective, we explain what it means and enumerate potential use cases.

\paragraph{Inter-hearing Analysis}
An inter-hearing analysis examines linguistic trends across hearings over time and committees, correlating them with societal trends to see how legislative discussions reflect or influence broader shifts. By analyzing speech styles and topic frequencies, we can track changes in communication strategies and focus areas between parties. Text descriptors like word diversity, homogeneity, and readability indexes, as shown by \citet{tucker2020data}, reveal significant shifts in lawmakers' speech styles. Because congressional hearings are an essential step for lawmakers to decide on future policies, hearings could also serve as indicators of legislative directions. \citet{wischnewsky2021financial} has shown this approach in the financial domain by constructing an indicator from testimonies of Federal Reserve chairmen and comparing it to the U.S.'s financial stability.

\paragraph{Intra-hearing Analysis}

Analyzing speaker dynamics within a single hearing can provide insights into power relationships based on seniority, party affiliation, or gender. Within a hearing, various types of exchanges occur, including question-and-answer sessions, debates, objections, and presentations of evidence. Examining these interactions can reveal the power dynamics and relationships among participants. Additionally, tracking topic shifts within a hearing can show how these changes influence the final decisions. The nature and flow of congressional exchanges, such as rebuttals and public comments, reflect the continuity of discourse and can impact the formulation of policies.

\paragraph{Participant-level Analysis}
Additionally, the analysis could be performed on each hearing participant, including both congress members and witnesses. With data on congress members' political affiliation and per-sentence linguistic properties, one could uncover partisan speech style differences. For instance, \citet{bayram2019s} demonstrated that a speaker's political party could be predicted based on word choice, indicating that partisanship can be inferred from simple semantic properties. Moreover, a particular participant's voting behavior can be predicted based on his legislative speech, as shown in \citet{budhwar2018predicting}.

\begin{figure*}[ht]
  \includegraphics[width=\textwidth]{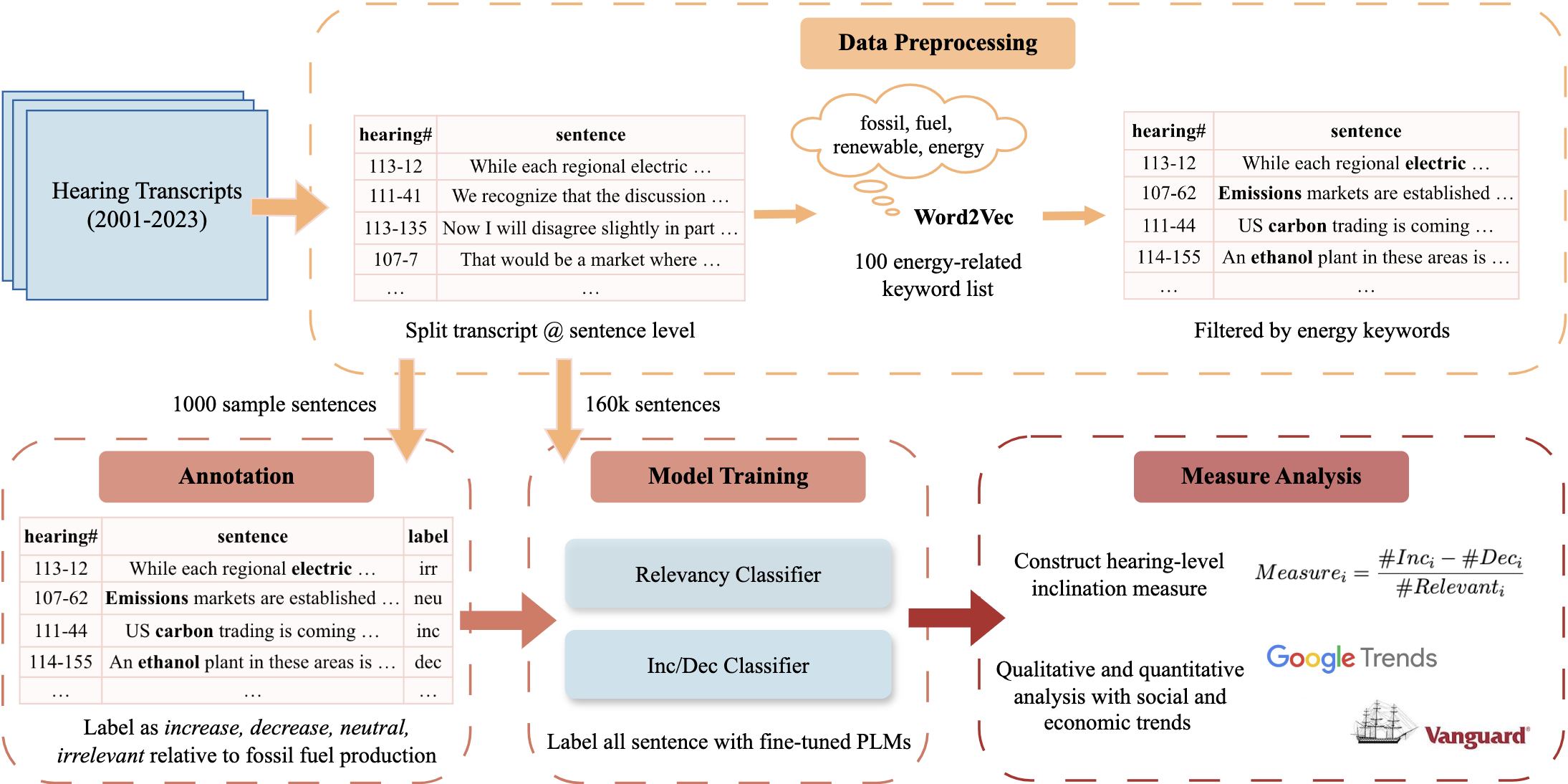}
  \caption{Fossil fuel sentiment analysis workflow.}
  \label{fig:pipeline}
\end{figure*}
\subsection{Quantifying Energy Policy Inclination}

We present a case study using our CoCoHD dataset to analyze the energy industry within the context of climate change. Congressional hearings shape energy sector strategies, providing valuable perspectives to stakeholders such as energy firms, environmental associations, and investors. However, manually analyzing hearings to gauge attitudes toward fossil fuels and clean energy is labor-intensive and subjective. Thus, employing the CoCoHD dataset, we streamline the inter and intra-hearing inclination analysis to understand the committee's stance on fossil fuels and clean energy. We detail our analysis pipeline in Figure~\ref{fig:pipeline}. 

\paragraph{Data Preprocessing}
We investigate the Energy and Commerce Committee due to its direct impact on the energy sector. The CoCoHD dataset already offers baseline accessibility so we only need to conduct minimal preprocessing to tailor the dataset for our use case. From a pool of 1,604 hearing transcripts, we exclude a subset of files that contain manual record errors, identified by eight types of missing text data markers. This results in a dataset comprising 1,586 hearing transcripts, spanning from 2001 to 2023. To support sentence-level semantic analysis, we segmented the text into valid sentences and standardized them into a clean format, resulting in over two million sentences.

Before analyzing the semantic nuances of hearing transcripts, we conduct a high-level analysis of the Energy and Commerce Committee to understand its structure, composition, and discussion focus over time. We categorize the committee into six subcommittees, excluding 3.6\% of hearings without affiliation. 

During the analysis process, we also created word clouds based on hearing titles and observed a temporal shift in focus: from health and legislative amendments in the early 2000s to energy discussions in the early 2010s, and clean energy and sustainability in the 2020s. These trends highlight the need to filter out irrelevant noise for effective analysis of energy usage dynamics.

Inspired by these findings, we implement keyword filtering to identify sentences that may be of relevance for further analysis. We start with four core terms: fossil, fuel, renewable, and energy. Using Word2Vec similarity scores, we expand this list to the top 100 words relevant to energy concepts. Filtering sentences with at least one energy keyword, we derive a set of around 160K sentences.

\paragraph{Manual Annotation}
To discern supportive and discouraging stances on energy usage, we manually annotated semantic inclinations of sentences based on energy industry context due to the absence of labeled data. We randomly selected 50 sentences from each year between 2001 and 2020, resulting in a dataset of 1,000 sentences from the Energy and Commerce Committee hearings. Transcripts for 2021 and 2022 were unavailable due to the prioritization of video records over textual records during the pandemic, and only limited transcripts for 2023 were accessible at the time of data collection.

We clarify that the fossil fuel industry comprises of US-based companies engaged in coal, oil, diesel, and natural gas exploration, production, and utilization, whereas renewable energy encompasses solar, hydro, hydrogen, wind, nuclear, and biofuel sectors. The labels are then defined based on fossil fuel production and usage. Sentences were classified into four categories: increase (p), decrease (d), neutral (n), and irrelevant (i). Increased sentences are those indicating support for fossil fuel production and usage, evidenced by direct pro-fossil fuel statements or indirect stances against clean gas or biofuel. Decrease sentences, on the other hand, indicate discouragement toward fossil fuel production and usage. Neutral sentences encompass mixed or conflicting preferences towards either source or those that do not express a clear opinion. Despite an initial filtering based on energy-related keywords, we encountered many sentences mentioning energy-related terms in contexts unrelated to energy policies or resources, or simply discussing legislative procedures involving committee and position names. Hence, we introduce a fourth category as irrelevant sentences that lie outside the scope of our analytical interest.

The labeling process was conducted by three coauthors with a foundational background in finance. Each sentence is first labeled independently by two annotators. In cases of conflicting labels, a third person reviewed and resolved the discrepancies. We detail the annotation guide(s) in Appendix~\ref{sec:annotation_guide_ap}. In summary, out of the 1,000 labeled sentences, there are 133 increase sentences, 183 decrease sentences, 393 neutral sentences, and 291 irrelevant sentences.

\paragraph{Model Training}

We train two classifiers using pre-trained language models (PLMs) to generalize the four labeling heuristics to the entire dataset. The relevancy classifier is trained on 1,000 sentences labeled as relevant or irrelevant, including all "increase," "decrease," and "neutral" sentences. The increase-decrease classifier is trained on the 709 relevant sentences, covering the "increase," "decrease," and "neutral" categories. We fine-tune RoBERTa and RoBERTa-large models, performing a grid search for optimal hyperparameters. Results indicate that the relevancy classifier outperforms the increase-decrease classifier, with RoBERTa-large showing superior performance in both tasks.

\paragraph{Inclination Measure Construction} We leverage labeled sentences within each hearing transcript to calculate a document-level inclination measure regarding the hearing's stance on pro-fossil fuel and pro-clean energy usage and production. For document $i$, we employ the following formula:
\[ Measure_i = \frac{\#Inc_i - \#Dec_i}{\#Relevant_i} * 100\%\]
where $\#Inc_i$ is the number of increase sentences in document $i$, $\#Dec_i$ is the number of decrease sentences in document $i$, and $\#Relevant_i$ is the total number of relevant sentences in document $i$. The resulting $Measure_i$ serves as a document-level metric, where a positive value signifies a pro-fossil fuel stance and a negative value indicates a pro-clean energy stance. In cases where there are equal numbers of increase and decrease sentences, or if the document lacks sentences labeled as increase or decrease, $Measure_i$ can be zero.

\paragraph{Market Analysis}To validate our proposed inclination measure, we perform a qualitative assessment by comparing its time series with Google Trends, based on two search key terms: "clean energy" and "fossil fuel".\footnote{\url{https://trends.google.com/trends/}} The detailed analysis on both of these can be found in Appendix~\ref{sec:gt_appendix}.

\paragraph{Regression Analysis}To demonstrate the economic significance of our inclination measure in the energy market, we conducted a quantitative analysis using the Vanguard Energy Index Fund ETF (VDE), which monitors the performance of the US energy sector.\footnote{\url{https://investor.vanguard.com/investment-products/etfs/profile/vde}} The VDE fund encompasses companies within specific energy sub-sectors such as fossil fuel and oil and gas services while excluding renewable and nuclear energy sources.

It is important to note that not every hearing may have a significant impact on the energy market. From a market perspective, those held during periods of high momentum or uncertainty are more likely to have substantial effects. To filter our dataset and identify hearings of importance, we implemented three layers of filtering. First, we selected hearings containing at least 22 relevant sentences using our prior keyword filtering methodology, representing the 70th percentile of relevancy counts. Next, we applied additional filters based on the previous seven days' market volatility and the relative strength index (RSI). Specifically, we filtered for hearings that occurred when market volatility was at or above 0.012 (representing the 50th percentile of VDE volatility) and when the RSI was outside the 30 to 70 range, a common threshold for RSI evaluation.

\begin{table}[ht]
\centering
\footnotesize
\begin{tabular}{ccc}
\toprule
\textbf{Dependent Variable} & $\alpha$ & $M (\beta_1)$\\
\midrule
Return\_5 & 0.004 & 0.0005* \\ 
Return\_6 & 0.0039 & 0.0007** \\ 
Return\_7 & 0.0059 & 0.0005 \\ 
\midrule
Vol\_7 & 0.2744*** & -0.0019** \\
Vol\_14 & 0.2818*** & -0.0015** \\
Vol\_21 & 0.2812*** & -0.0016** \\
Vol\_28 & 0.2791*** & -0.002*** \\ 
\bottomrule
\end{tabular}
\caption{Linear regression analysis results for market analysis. Here, M represents the inclination measure and $\alpha$ represents the intercept. Significance levels are denoted as ***$p<0.01$, **$p<0.05$, *$p<0.1$.}
\label{tab:linear_regression}
\end{table}

To quantitatively assess the impact of significant congressional hearings on the energy market, we conduct linear regression using our measure of inclination as the independent variable. Table~\ref{tab:linear_regression} illustrates the relationship between our inclination measure and the 5, 6, and 7-day returns, as well as the 7, 14, 21, and 28-day volatility following these congressional hearings. A statistically significant positive coefficient for the 5 and 6-day returns suggests that a pro-fossil fuel hearing stance correlates with an upward movement in the energy market in the subsequent days, whereas negative sentiment corresponds to a downturn.

Moreover, we observe a reduction in market volatility for up to a month following these hearings. This indicates that discussions during the hearings contribute to greater clarity within the energy markets. Lower market volatility signifies increased certainty and consensus among market participants. This finding is particularly noteworthy given that our analysis focused on hearings preceded by unusually high volatility.

In summary, our inclination measure shows a significant correlation with market returns and volatility, highlighting its potential in predicting market trends of congressional stances on energy policies.

\section{Related Work}

\paragraph{NLP in Congressional Analysis} 
Many studies have demonstrated the potential of NLP to improve our understanding of the legislative system. Congress has been the subject of many quantitative text analyses to extract political, financial, and socioeconomic insights. Politically, researchers have examined partisanship, ideology, and member connections. For instance, \citet{tucker2020data} leveraged a data science approach to analyze 138 years of congressional speeches, revealing trends in political speech complexity, sentiment, and partisan differences. Their analysis demonstrated two key trends: First, congressional hearings became more readable over time but have experienced a sharp decline in readability since the 1970s. Second, there has been an increase in polarized statements, with more instances of both highly positive and highly negative language. These trends suggest that the nature of congressional communication has shifted, potentially reflecting more complex or specialized legislative language, alongside increasingly divided political rhetoric, which may mirror the growing polarization in society and within the legislative body itself. \citet{bayram2019s} demonstrated that a speaker's political party could be predicted based on word choice, indicating that partisanship can be inferred from simple semantic properties. More specifically, \citet{diermeier2012language} highlighted the role of cultural references in congressional speeches in distinguishing the political ideologies of congress members. Beyond the US Congress, \citet{lima2023analysis} utilized NLP and machine learning to analyze speeches in the Brazilian National Congress, employing network analysis to assess relationships between members, which in turn helps understand party cohesion. Their findings suggest that similar methods could be applied to uncover hidden power dynamics or biases in other legislative bodies.

\paragraph{Social Dynamics in Congressional Hearings} 
Researchers have explored the social dynamics in congressional hearings. \citet{bisbee2022yellin} found evidence of gender bias in congressional hearings by studying language properties, demonstrating how language can subtly reinforce gender hierarchies even in formal settings. \citet{ban2022does} investigated how the presence of women in congressional committees impacted discussion dynamics, with a shift in norms toward more in-depth exchange. Their research underscores the importance of representation, showing that increased diversity can challenge existing norms and lead to more inclusive legislative discourse. 

\paragraph{Financial Sentiment in Congressional Hearings} 
From the financial perspective, \citet{wischnewsky2021financial} analyzed Fed Chair testimonies in congressional hearings, finding that speeches expressing concerns about financial stability influenced US monetary policy. Their research demonstrated that when the Federal Reserve Chair emphasized financial stability in discussions with Congress, the Fed adjusted its monetary policy accordingly. Negative sentiment around financial stability had a stronger influence, leading to a more accommodative policy than traditional models would suggest. This indicates the Fed's preference for responding to financial instability rather than acting preemptively, consistent with remarks from officials like Greenspan and Bernanke.

Other related works are discussed in Appendix \ref{ap:other-related-work}. These datasets, especially congressional hearing datasets, although mostly comprehensive, are often tailored to specific research objectives, making them less adaptable for exploring diverse research inquiries. This, once again, underscores the necessity for a unified congressional hearing dataset to address this limitation.

\section{Conclusion}
We introduce CoCoHD, a comprehensive U.S. Congressional hearings dataset with over 32,000 transcripts and metadata from 1997 to 2024. Our case study on the Energy and Commerce Committee demonstrates CoCoHD's utility by analyzing congressional attitudes toward fossil fuels and clean energy. We developed an inclination measure per hearing, quantifying stances on fossil fuels versus clean energy, validated through Google search trends and a statistically significant correlation with energy market indicators. 

CoCoHD offers extensive opportunities for researchers to explore congressional perspectives on various critical social issues. By analyzing these topics, researchers can gain insights into legislative priorities, ideological leanings, and evolving policy discussions within Congress.

\newpage

\section*{Limitations}
During our annotation process, labeling a single sentence was extremely complex due to the economic, financial, and political knowledge required. The vast breadth of topics necessitated constant research, making crowd-sourced annotations infeasible. As a result, we manually annotated 1,000 sentences. Despite covering many topics, each had very few related sentences, making it challenging for the language model to learn a consistent pattern. We believe that currently annotated sentences are insufficient to capture all sentiments toward fossil fuels and clean energy. Future work should aim for a more simplistic and systematic annotation method.

Due to the lack of a systematic method to study congressional hearings' stance on various topics and the laborious annotation process, we have exclusively focused on demonstrating CoCoHD's effectiveness in understanding trends in energy sources. Consequently, it remains unclear how much insight congressional hearings by other committees provide into other social topics.

Other limitations include not exploring diverse uses of the data, such as different tasks and congressional committees. More in-depth NLP tasks, such as aggregating political data and studying the flow of conversations at higher granularity, should be considered. Future research could benefit from using the metadata of member lists we provided to enhance the analysis.

\section*{Ethics Statement}
Our work adheres to ethical considerations, although we acknowledge certain biases and limitations in our study. We do not identify any potential risks stemming from our research; however, we recognize the presence of geographic and gender biases in our analysis.

\paragraph{Geographic Bias} In conducting research and reviewing related literature on congressional hearings and legislative processes, several ethical considerations and potential biases must be acknowledged, as they have significant implications for both the validity of the findings and the broader impact on public understanding and policy. One prominent concern is geographic bias, which arises from focusing exclusively on U.S. congressional hearings. This narrow focus may limit the generalizability of the research to legislative systems worldwide, as political structures, cultural norms, and communication styles can vary significantly across countries. By not incorporating data from other legislative bodies, the insights drawn from U.S. congressional hearings may reflect uniquely American political dynamics, such as its two-party system, federalism, and specific socio-political issues, and fail to capture the legislative nuances present in more pluralistic or parliamentary systems elsewhere. This raises concerns about the applicability of such findings to global contexts, as they might lead to skewed interpretations when applied to different governmental structures, hindering comparative legislative studies.

\paragraph{Gender Bias} Gender bias is another critical issue, stemming from the historical over-representation of male members in the U.S. Congress. This imbalance can lead to findings that disproportionately reflect male-dominated viewpoints, possibly marginalizing issues that are more relevant to women or gender minorities. Additionally, linguistic patterns and rhetorical styles that are more commonly associated with male speech could skew analyses, leading to conclusions that do not fully capture the diversity of communication present in a truly representative legislative body. This can perpetuate existing gender biases in policy analysis and decision-making, reinforcing unequal power dynamics in political discourse.

Addressing these biases is essential for ensuring that the research provides a more accurate, inclusive, and globally relevant understanding of legislative processes. Expanding the scope of analysis to include a more diverse range of legislative systems and critically engaging with political discourse are vital steps toward achieving more comprehensive and equitable research outcomes.

\paragraph{Annotation Ethics} All annotations were performed by the authors, ensuring that no additional ethical concerns arise from the annotation process. 

\paragraph{Publicly Available Data} We specify the datasets that will be made publicly available and indicate the applicable licenses under which they will be shared.

\section*{Acknowledgements}
We appreciate the generous support of Azure credits from Microsoft made available for this research via the Georgia Institute of Technology Cloud Hub.

\newpage

\bibliography{anthology,custom}
\bibliographystyle{acl_natbib}

\appendix

\section{Identifying SubCommittees}
\label{sec:identify_subcommittees_ap}
The process begins by iterating through all files in a directory. For each file, the content is read into a single string. Using regular expressions, the text is scanned for instances of "SUBCOMMITTEE ON" or "Subcommittee on," followed by capital letters, with any whitespace characters (including newlines) allowed between words. Two patterns are used to match these instances: one for uppercase matches and another for lowercase matches.

The matches found are then combined and processed. Each match is cleaned by replacing multiple whitespace characters with a single space and removing any leading or trailing whitespace. If the cleaned list of matches is not empty, the matches are further processed to ensure they are formatted consistently: all entries are converted to title case and any duplicates are removed.

If there are multiple unique subcommittee names found, they are added to a list. If no matches are found in a file, the file name is added to a separate list to keep track of files without any subcommittee information. If only a single subcommittee name is found, it is added to the list of found subcommittees as is.

This approach ensures that subcommittee names are accurately extracted, cleaned, and compiled from each file, providing a comprehensive list of subcommittees and highlighting any files that do not contain subcommittee information.

\section{Identifying Hearing Contents}
\label{sec:hearing_contents_ap}

We define the process of identifying the Contents subsection in congressional hearings: 

A function is defined to extract and process the contents from the given file path. The function reads the file content into a string and searches for the line containing "CONTENTS:" (case insensitive) using the regex pattern: 

\texttt{C\textbackslash s*O\textbackslash s*N\textbackslash s*T\textbackslash s*E\textbackslash s*N\textbackslash s*T\textbackslash s*S}. A regular expression is used to locate the "CONTENTS:" line, allowing for any number of spaces between the letters.

Next, the pattern to stop extraction is defined to match a line ending with dots followed by spaces and numbers (e.g., ".    67") using the regex: \texttt{\textbackslash.+\textbackslash s\{4,\}\textbackslash d+\textbackslash n}
. The content between the "CONTENTS:" line and the last match of the end pattern is extracted.

A function is then defined to clean the extracted text by performing several operations: removing unnecessary newline characters using a regex pattern, removing the word "Page" using the regex pattern \texttt{\textbackslash bPage\textbackslash b}, stripping leading and trailing whitespace, replacing sequences of dots followed by spaces with a single space using the regex pattern \texttt{\textbackslash.\{2,\}\textbackslash s*}
, removing sequences of dashes using the regex pattern \texttt{-\{2,\}}, and replacing multiple spaces with a single space using the regex pattern \texttt{\textbackslash s\{2,\}}
.

The cleaned text is then split into a list of strings, and the word "Witnesses" is removed from this list. Finally, the cleaned list of contents is returned.

\section{Identifying Prepared Statements}
\label{sec:prepared_statements_ap}

We define the process of identifying the Prepared Statements as well as Opening Statements in Congressional Hearings: 

To identify prepared statements as well as opening statements from a text file, the process begins by reading the entire content of the file. The content is then scanned to find lines that contain the phrase "STATEMENT OF," and the text is split at each occurrence of this phrase. Any resulting empty segments from this split are removed to ensure only meaningful text remains.

Next, the split text segments are organized into pairs, where the first line of each segment becomes a key, and the rest of the segment is treated as its value. These pairs are stored in a dictionary for further processing.

The values in this dictionary are then checked for the presence of a backslash. If a backslash is found, the key and value are updated: the part of the value before the backslash is added to the key, and the part after the backslash becomes the new value. This adjustment ensures that the keys and values are more accurately represented.

Further refinement of the keys involves identifying certain split characters (like backslashes or commas) within the keys. If such characters are found, the keys are split at these points, and the resulting segments are used to update the keys and values appropriately. This step ensures that the keys are properly formatted and the values are correctly associated with them.

The next phase involves identifying segments of text that contain the phrase "Prepared Statement of." When such segments are found, the dictionary is updated to separate the part before this phrase from the part that includes and follows it. This step creates new key-value pairs for each identified "Prepared Statement of" segment.

This process is repeated iteratively until no further changes occur, ensuring all relevant segments are correctly identified and processed. The dictionary is then refined by adjusting the keys and values based on specific formatting patterns, ensuring that the keys are correctly structured.

Finally, the keys in the dictionary are converted to title cases, making them more readable. This involves capitalizing the first letter of each word while keeping the rest of the letters lowercase. This transformation is applied to all keys and any nested dictionaries within the main dictionary. The final result is a well-organized list of key-value pairs that clearly identifies and distinguishes between prepared statements and opening statements. 

\begin{table*}[ht]
\centering
\footnotesize
\begin{tabular}{cl}
\toprule
\textbf{Label}             & \multicolumn{1}{c}{\textbf{Rule}}                                                                                                                                                                                                                                      \\ \midrule
\begin{tabular}[c]{@{}c@{}} Increase \\ (fossil fuel usage)\end{tabular} & \begin{tabular}[c]{@{}l@{}}1. Sustained use of fossil fuel\\ 2. Against gas or biofuel mix\end{tabular}                                                                                                                                                                  \\ \midrule
\begin{tabular}[c]{@{}c@{}} Decrease \\ (fossil fuel usage) \end{tabular}  & \begin{tabular}[c]{@{}l@{}}1. Reduction in the amount of fossil fuel used\\ 2. In favor of reducing carbon emission without being more specific\\ 3. For gas or biofuel mix, e.g. ethanol\end{tabular}                                                                   \\ \midrule
Neutral                    & \begin{tabular}[c]{@{}l@{}}1. Conflict presentation of both inclination\\ 2. Relevant to energy but doesn't talk about increase/decrease usage\\ 3. Quote of someone without own opinion\\ 4. Maintaining the current state of oil or clean energy industry \\ 5. Diversification of energy usage \end{tabular} \\ \midrule
Irrelevant                 & \begin{tabular}[c]{@{}l@{}}1. Subject discussed unrelated to energy or unclear\\ 2. Questions except rhetorical ones\\ 3. About electric grid without mentioning a specific energy resource\\ 4. Cannot be clearly labeled with the other ones\end{tabular}              \\ \bottomrule
\end{tabular}
\caption{General annotation rules.}\label{tab:1}
\end{table*}

\begin{table*}[ht]
\footnotesize
\centering
\begin{tabular}{cll}
\toprule
\textbf{Category} & \multicolumn{1}{c}{\textbf{Increase}}  & \multicolumn{1}{c}{\textbf{Decrease}}  \\ \midrule Policy  & \begin{tabular}[c]{@{}l@{}}1. In favor of cleaner fossil fuel exploration, \\ production or usage\\ 2. Against using more renewable energy\end{tabular} & \begin{tabular}[c]{@{}l@{}}1. Against fossil fuel exploration, \\ production or usage\\ 2. Against cleaner fossil fuel exploration, \\ production or usage\\ 3. In favor of using more renewable energy\end{tabular} \\ \midrule
Environment      & \begin{tabular}[c]{@{}l@{}} 1. Negative environmental impact of \\ renewable energy industry   \end{tabular}  & \begin{tabular}[c]{@{}l@{}} 1. Negative environmental impact of \\ fossil fuel industry  \end{tabular}   \\ \midrule
Society           & \begin{tabular}[c]{@{}l@{}} 1. Economic significance of fossil fuel \\ industry to certain regions \end{tabular}  & \begin{tabular}[c]{@{}l@{}} 1. Negative impact of fossil fuel \\ industry on various aspects of society  \end{tabular}   \\ \midrule
Market            & \begin{tabular}[c]{@{}l@{}}1. Fossil fuel market is competitive\\ 2. Fossil fuel market is growing\end{tabular}                                & \begin{tabular}[c]{@{}l@{}}1. Fossil fuel market is not competitive\\ 2. Fossil fuel market is shrinking\end{tabular}                                                                                 \\ \midrule
Trade             & \begin{tabular}[c]{@{}l@{}}1. In favor of fossil fuel export\\ 2. Against fossil fuel import\end{tabular}                                            & \begin{tabular}[c]{@{}l@{}}1. Against fossil fuel export\\ 2. In favor of fossil fuel import\end{tabular}                                                                                                      \\ \bottomrule
\end{tabular}
\caption{Category-specific annotation rules.}\label{tab:2}
\end{table*}

\section{Identifying Active Speakers}
\label{sec:active_speakers_ap}
We define the process of identifying members of the hearing who have spoken in the transcript: 

To identify repeated starting sentences in paragraphs, the process begins by reading the content of the text file. The text is split into individual sentences, ensuring that sentence endings are correctly identified while ignoring periods in abbreviations. Any leading or trailing whitespace is removed from these sentences.

Next, the text is split into paragraphs. Each paragraph is trimmed of any leading or trailing whitespace, and only non-empty paragraphs are considered.

The sentences are then analyzed to count how many times each one appears in the text. Sentences that appear more than once are identified as repeated sentences.

For the paragraphs, a pattern is used to match common titles followed by full names (such as "Chairwoman", "Secretary", "Mr.", "Ms.", "Mrs.", and "Dr."). If a paragraph starts with one of these titles, the first sentence of the paragraph is extracted. If this first sentence is among the repeated sentences, it is counted as a repeated starting sentence.

To further refine the results, any sentences that consist only of incomplete titles (e.g., "Mr.", "Mrs.", "Dr.", "Ms.", "Chairwoman", "Secretary") are filtered out. This ensures that only meaningful repeated starting sentences are considered.

Finally, the repeated starting sentences are printed along with the count of how many times each one appears at the start of paragraphs. The process returns a list of these repeated starting sentences, providing a clear view of common patterns in the text.

\section{Annotation Guide(s) For Energy Case Study}
\label{sec:annotation_guide_ap}
The annotation guide comprised a general rule section in Table~\ref{tab:1} for baseline guidelines and a category-specific section in Table~\ref{tab:2} providing detailed instructions, categorizing target sentences into topics of policy, environment, society, market, and trade. 

\begin{table*}[ht]
\centering
\footnotesize
\begin{tabular}{cccccc}
\toprule
\textbf{Classifier}        & \textbf{Model} & \textbf{F1 Mean} & \textbf{F1 Std} & \textbf{LR} & \textbf{BS} \\ \midrule
\multirow{2}{*}{relevancy} & RoBERTa-base        & 0.878            & 0.038           & 1e-5        & 4              \\
                           & RoBERTa-large  & \textbf{0.886}   & \textbf{0.009}           & 1e-6        & 4              \\ \midrule
\multirow{2}{*}{inc/dec}   & RoBERTa-base        & \textbf{0.620}   & 0.048           & 1e-5        & 4              \\
                           & RoBERTa-large  & 0.618            & \textbf{0.019}           & 1e-5        & 32             \\ 
                           \bottomrule
\end{tabular}
\caption{Here LR stands for the best learning rate and BS stands for the optimal batch size. The F1 mean and standard deviation are calculated as an average of 3 seeds used for all models. The F1 Mean score and the F1 standard deviation of the top-performing model for each task are indicated in bold.
}
\label{tab:3}
\end{table*}

\begin{figure*}[ht]
    \includegraphics[width = \textwidth]{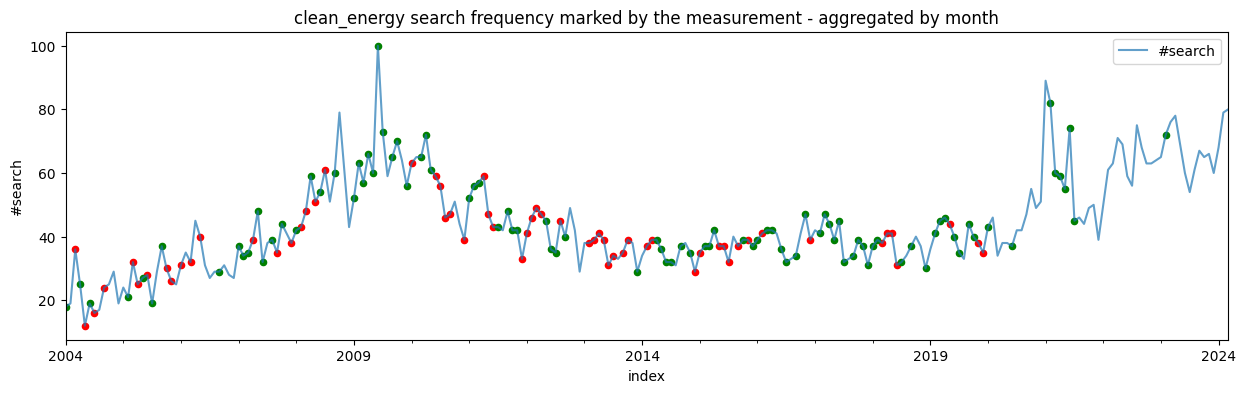}
    \caption{The blue line shows the search frequency trend, with red dots indicating a positive metrics score (increased fossil fuel consumption) and green dots indicating a negative score (preference for clean energy). Months without dots had no hearings, and dates beyond 2020 lack sufficient transcript data.}
    \label{fig:clean_enery}
\end{figure*}

\section{Detailed Model Training Process}

\label{sec:model_training_ap}

To generalize the four labels to the entire dataset, we propose training two classifiers using pre-trained language models (PLMs). The relevancy classifier is trained on 1,000 sentences labeled as relevant or irrelevant, with all "increase," "decrease," and "neutral" sentences labeled as relevant. The increase-decrease classifier is then trained on the 709 relevant sentences, encompassing the "increase," "decrease," and "neutral" categories. We fine-tune two models, RoBERTa and RoBERTa-large \cite{liu2019roberta}. To identify optimal hyperparameters, we perform a grid search across four learning rates (1e-4, 1e-5, 1e-6, 1e-7) and four batch sizes (4, 8, 16, 32). During training, we employ three different random seeds (5768, 78516, 944601) and calculate the average weighted F1 scores. The results for both classifiers and models are presented in Table~\ref{tab:3}.

We observe that the relevancy classifier outperforms the increase-decrease classifier, likely due to the former being a binary classification task, whereas the latter is a harder three-class task. The "increase," "decrease," and "neutral" categories in the increase-decrease task involve more semantic nuances, making accurate classification more challenging. This observation aligns with our manual data labeling experience, where most conflicting annotations occurred in the increase-decrease labeling stage. Additionally, RoBERTa-large demonstrates superior performance compared to the baseline RoBERTa model in both tasks, exhibiting higher mean F1 scores and a narrower F1 standard deviation. Consequently, we utilize RoBERTa-large to annotate all the filtered sentences.

\begin{figure*}[ht]
    \includegraphics[width = \textwidth]{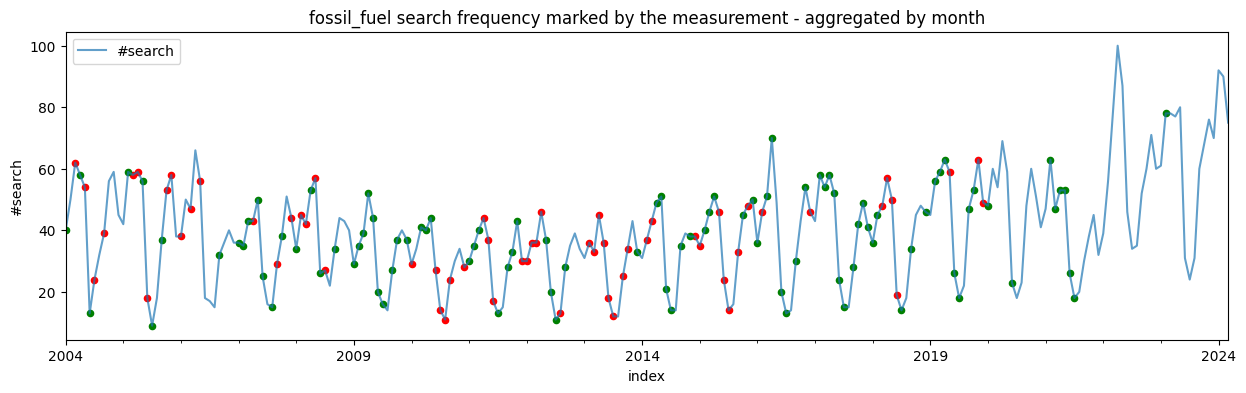}
    \caption{The blue line shows the search frequency trend, with red dots indicating a positive metrics score (increased fossil fuel consumption) and green dots indicating a negative score (preference for clean energy). Months without dots had no hearings, and dates beyond 2020 lack sufficient transcript data.}
    \label{fig:fossil_fuels}
\end{figure*}

\section{Google Trends Analysis of Congressional Hearings
Inclination to Fossil Fuels}
\label{sec:gt_appendix}

As shown in Figure \ref{fig:clean_enery}, the keyword "clean energy" peaks between 2009 and 2010, aligning with consistently negative scores in our monthly metrics. This period corresponds to the introduction of the Clean Energy Act in 2009, which set renewable energy standards, promoted energy efficiency, and incentivized clean energy projects\footnote{\url{https://www.congress.gov/bill/111th-congress/house-bill/2454}}. This explains the sustained interest in clean energy that year, as reflected in our analysis, validating our measure's predictive capability. After this, there is a cluster of pro-fossil fuel stances until around 2016, followed by a shift towards clean energy discussions post-2016. This transition aligns with advancements in clean energy technologies and heightened climate change awareness, marked by the Paris Agreement in 2015\footnote{\url{https://www.un.org/en/climatechange/paris-agreement}}. Compared to Google search trends, our measure more clearly highlights this shift, underscoring its value in predicting societal trends.

As shown in Figure \ref{fig:fossil_fuels}, the frequency of searches for "fossil fuel" exhibits periodic fluctuations, likely due to seasonal patterns such as increased reliance on fossil fuels during winter for heating. Although our inclination measure does not capture this seasonal variation, as seasonality is not a significant factor in congressional interest in fossil fuels and clean energy, we note a correlation where increased interest in clean energy coincides with heightened searches for fossil fuels. This correlation is evident from the peak in searches around 2016, coinciding with the tabling of the Paris Climate Agreement, and the sustained attention on fossil fuels post-2016. This pattern demonstrates the ongoing debate and shifting dynamics between fossil fuels and clean energy initiatives.

\section*{Additional Related Work}
\label{ap:other-related-work}
\paragraph{NLP in Other Legislative Systems} 
Apart from Congress, NLP has also demonstrated its informative value in other legislative branches. One outstanding use case is to predict vote outcomes. \citet{budhwar2018predicting} predicted vote outcomes based on verbal utterances during the legislative process from elected representatives. This raises the potential for biases in how certain speech patterns or phrasing might skew predictions toward specific political ideologies, possibly influencing public opinion. \citet{korn2020deep} predicted congressional roll call votes from legislative texts using a novel deep learning model. Another category of task that researchers have used NLP to tackle is legislation analysis and processing. For example, \citet{van2001modeling} explored the translation of legislation from a natural language to a formal language like UML/OCL using NLP techniques. \citet{spinosa2009nlp} contributed by presenting a metadata-oriented approach to the consolidation of legislative texts, utilizing NLP techniques and XML-based standards for metadata annotation. 

\paragraph{Legislative Text Datasets} 
Some related work has shown advancements in the creation of Legislative Text Datasets. \citet{kornilova2019billsum} introduced BillSum, the first dataset for summarization of US Congressional and California state bills, along with benchmarked extractive methods. \citet{libgober2024comprehensive} proposed a dataset containing 49,746 laws spanning from 1789 to 2022 by aggregating a complicated patchwork of documents published in numerous and inconsistent formats.  For congressional discourse, \citet{thomas2006get} developed a congressional speech corpus with labels indicating a speaker's stance on debated legislation, facilitating the study of conversation flow in congressional debates. Additionally, \citet{maher2020social} presented a dataset to examine the evolution of social scientists' impact in congressional hearings, documenting the frequency at which scientists such as anthropologists, economists, political scientists, and sociologists appeared before hearing sessions. 

These datasets, especially congressional hearing datasets, although mostly comprehensive, are often tailored to specific research objectives, making them less adaptable for exploring diverse research inquiries. This, once again, underscores the necessity for a unified congressional hearing dataset to address this limitation.

\end{document}